\documentclass{article} 
\usepackage{iclr2027_conference,times}

\usepackage{amsmath,amsfonts,bm}

\def\eqref#1{equation~\ref{#1}}

\def\1{\bm{1}}

\DeclareMathAlphabet{\mathsfit}{\encodingdefault}{\sfdefault}{m}{sl}
\SetMathAlphabet{\mathsfit}{bold}{\encodingdefault}{\sfdefault}{bx}{n}

\usepackage{hyperref}
\usepackage{url}
\usepackage{multirow}
\usepackage{graphicx}
\usepackage{makecell}
\usepackage{subcaption}
\usepackage{wrapfig}
\usepackage{booktabs}
\usepackage{amsmath,amssymb}
\usepackage{longtable}
\usepackage{float}

\usepackage[most]{tcolorbox}
\usepackage{pgfplots}
\usetikzlibrary{pgfplots.groupplots}
\pgfplotsset{compat=1.3}
\usepackage{tikz}
\usetikzlibrary{patterns}

\newlength\savewidth\newcommand\shline{\noalign{\global\savewidth\arrayrulewidth
  \global\arrayrulewidth 1pt}\hline\noalign{\global\arrayrulewidth\savewidth}}

\newcommand\midline{\noalign{\global\savewidth\arrayrulewidth
  \global\arrayrulewidth 0.5pt}\hline\noalign{\global\arrayrulewidth\savewidth}}

\newcommand{\tablestyle}[2]{%
  \setlength{\tabcolsep}{#1}%
  \renewcommand{\arraystretch}{#2}%
  \centering
  \footnotesize
}
\definecolor{baselinecolor}{gray}{.9}

\definecolor{lightgreen}{RGB}{220,245,220}
\definecolor{lightred}{RGB}{245,220,220}
\definecolor{cornellred}{rgb}{0.7, 0.11, 0.11}
\definecolor{cadmiumgreen}{rgb}{0.0, 0.42, 0.24}
\definecolor{aliceblue}{rgb}{0.91, 0.94, 0.97}
\definecolor{darkblue}{rgb}{0.83, 0.89, 0.97}
\definecolor{Red7}{rgb}{0.941, 0.243, 0.243}
\definecolor{Green7}{RGB}{55, 178, 77}
\definecolor{Blue9}{rgb}{0.098,0.3,0.9}

\hypersetup{
  linkcolor = cornellred,
  citecolor  = cadmiumgreen,
  colorlinks = true,
  urlcolor = Blue9
}

\newcommand{\ie}{\emph{i.e.}}    
\newcommand{\eg}{\emph{e.g.}}    

\title{Qwen-Image-Flash: Rethinking the Training Recipe for Few-Step Distillation}

\author{%
  \begingroup
  \small
  \renewcommand{\arraystretch}{1.15}%
  \begin{tabular}{@{}l@{}}
    Tianhe Wu$^{1,2}$\thanks{Equal contribution.}, Zikai Zhou$^{3,2}$\footnotemark[1], Kun Yan$^{2}$, Kaiyuan Gao$^2$, Lihan Jiang$^2$, Jiahao Li$^2$, Jie Zhang$^2$, \\
    Ningyuan Tang$^2$, Shengming Yin$^2$, Xiaoyue Chen$^2$, Xiao Xu$^2$, Yilei Chen$^2$, Yuxiang Chen$^2$, \\
    Yan Shu$^2$, Yixian Xu$^2$, Yanran Zhang$^2$, Zihao Liu$^2$, Zhendong Wang$^2$, Zekai Zhang$^2$, Deqing Li$^2$, \\
    Liang Peng$^2$, Yi Wang$^2$, Zeke Xie$^3$, Jingren Zhou$^2$, Bo Zheng$^2$, Chenfei Wu$^{2}$\thanks{Corresponding author.} \\[0.2em]
    \normalfont\footnotesize
    $^1$City University of Hong Kong\;
    $^2$Alibaba, Qwen Image Group\\
    \normalfont\footnotesize
    $^3$The Hong Kong University of Science and Technology (Guangzhou)\\[0.2em]
    {\tt\small tianhewu-c@my.cityu.edu.hk}\; {\tt\small zzhou440@connect.hkust-gz.edu.cn}\\
    {\tt\small fulai.hr@alibaba-inc.com}\\
  \end{tabular}
  \endgroup
}

\iclrfinalcopy 

\begin{document}

\maketitle
\vspace{-4mm}

\begin{figure}[h]
  \centering
  \vspace{-5mm}
  \includegraphics[width=0.96\textwidth]{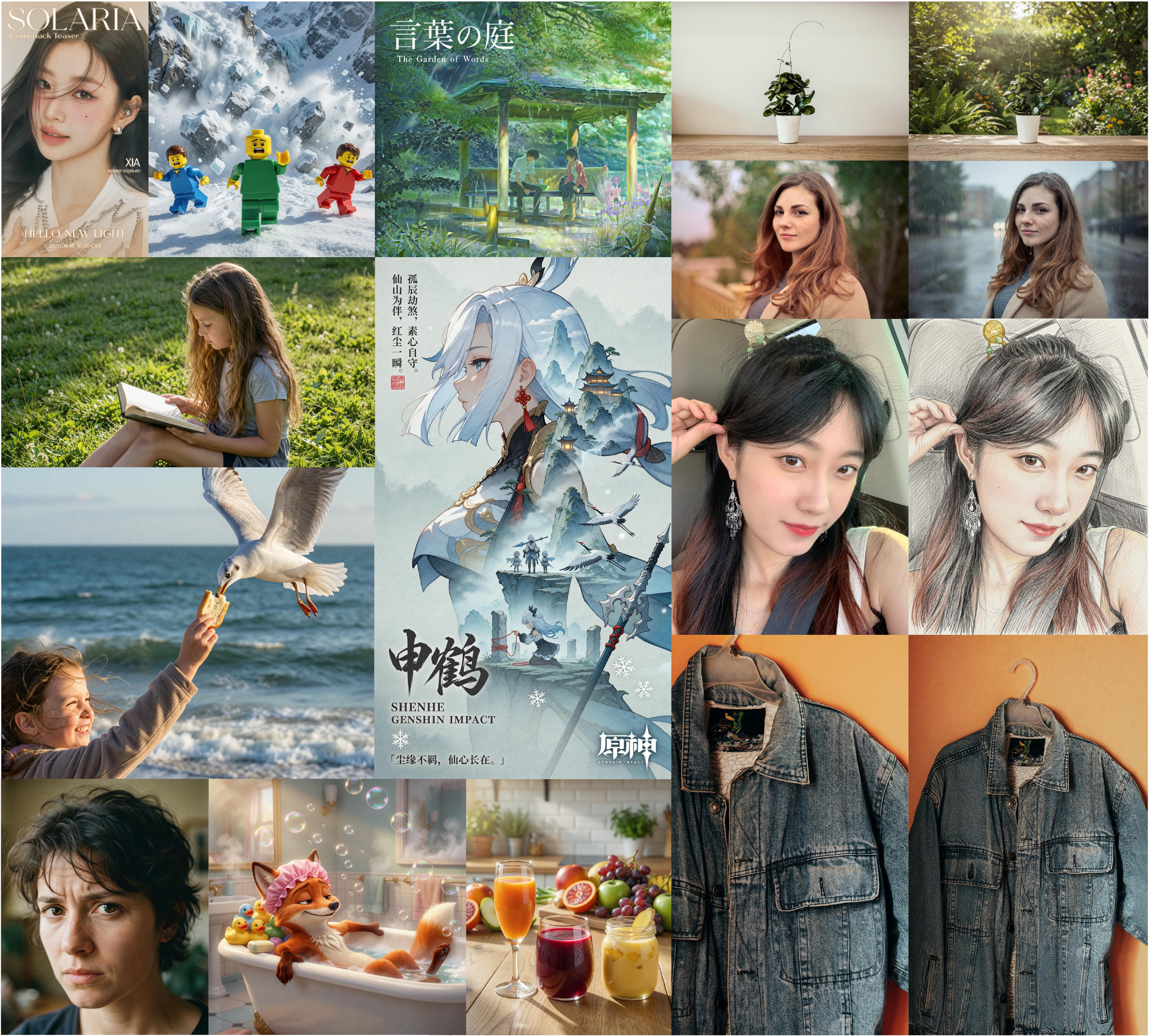}
  \caption{\textbf{Qualitative results of Qwen-Image-Flash.} T2I generation and image editing results with only $4$ NFEs, demonstrating unified few-step capabilities across generation and editing.}
  \label{fig:teaser}
\end{figure}

\begin{abstract}
Few-step distillation has emerged as a critical component in the development of advanced visual generative foundation models, substantially reducing inference overhead while enabling real-time generation and cost-efficient deployment across a broad range of practical scenarios. However, prior work has predominantly focused on advancing training objectives, while comparatively overlooking the \textit{training recipe}, which has become increasingly critical in the era of large-scale foundation models. In this work, we systematically revisit the training recipe under the well-established distribution matching distillation (DMD) framework for both text-to-image generation and image editing, focusing on three key dimensions: \textit{training data composition}, \textit{teacher guidance within DMD}, and \textit{task mixture}. Our empirical analysis reveals several non-obvious and counterintuitive phenomena, ultimately motivating the development of \textbf{Qwen-Image-Flash}. These findings highlight that effective few-step distillation depends not only on carefully designed objectives, but also on a principled training recipe.
\end{abstract}

\section{Introduction}
Visual generative models have emerged as a flexible and accessible paradigm for content creation, interactive editing, and multimodal applications~\citep{esser2024scaling,song2026awaking,liu2026ernieimagetechnicalreport,mao2026wan}. Built predominantly on diffusion~\citep{ho2020denoising,song2020score} and flow-based formulations~\citep{lipman2022flow}, modern foundation models can generate high-fidelity images from complex prompts, render dense structured text, and perform instruction-guided editing within a unified framework~\citep{zhao2026qwen}.

However, their practical deployment remains limited by iterative denoising or transport trajectories that require numerous function evaluations, resulting in substantial latency and computational cost. This sampling overhead is particularly prohibitive for latency-sensitive and resource-constrained scenarios, including interactive editing~\citep{meng2022sdedit}, on-device generation~\citep{li2023snapfusion,zhao2024mobilediffusion}, and large-scale content production~\citep{azuma1997survey,yin2025causvid}.

To alleviate this bottleneck, few-step distillation has emerged as a promising paradigm that compresses the generative dynamics of a multi-step teacher into a student capable of high-quality generation with only a small number of function evaluations (NFEs). Recent advances have largely centered on the design of distillation objectives, spanning trajectory-level alignment~\citep{geng2025mean}, consistency training~\citep{song2023consistency}, adversarial distillation~\citep{sauer2024adversarial}, and distribution matching~\citep{yin2024improved, yin2024one, jiang2025distribution, wu2026diversity}.

Nevertheless, a standard training recipe is not a safe bet: directly applying even widely adopted methods such as distribution matching distillation (DMD, \citealt{yin2024improved, yin2024one}) to large-scale visual generative models across broad and heterogeneous scenarios can lead to severe degradation or outright training collapse, as illustrated in Figure~\ref{fig:training_data}. This observation suggests that the \textit{distillation objective is only part of the story}: effective few-step distillation also hinges on the \textbf{training recipe} used to put it into practice.

Motivated by this gap, we use Qwen-Image-2.0~\citep{zhao2026qwen} as a representative large-scale visual generative foundation model to systematically investigate the DMD training recipe. For comprehensive evaluation, we establish two complementary benchmarks: T2I-Bench, with $1{,}800$ cases evenly distributed across three representative categories, and Editing-Bench, with $1{,}500$ challenging editing cases spanning six representative categories. Using this testbed, we examine three key dimensions: \textit{training data composition}, which defines the text-to-image (T2I) distillation distribution; \textit{teacher guidance within DMD}, which governs the incorporation of teachers with different capabilities; and \textit{task mixture}, which controls the joint optimization of T2I generation and image editing.

Our empirical analysis reveals three key findings. First, T2I distillation is highly sensitive to training data composition: neither greater data diversity nor stronger alignment with target scenarios necessarily improves performance, whereas semantically coherent data from a single category can exhibit surprisingly strong cross-category generalization. Second, incorporating post-trained teachers with enhanced capabilities into DMD is non-trivial, as naively using them throughout distillation can destabilize training and lead to severe structural collapse. We find that restricting post-trained teacher guidance to the final step effectively transfers their specialized capabilities while preserving the structural stability inherited from the base teacher. Third, in joint T2I-editing distillation, task mixture emerges as a critical factor, with a balanced T2I-to-editing training ratio yielding the strongest overall performance across both tasks.

Building on these findings, we develop \textbf{Qwen-Image-Flash}, a unified few-step model for both T2I generation and image editing. As illustrated in Figure~\ref{fig:teaser}, Qwen-Image-Flash achieves high-quality generation and editing with only $4$ NFEs, while retaining strong visual fidelity and versatile synthesis capabilities across diverse and complex scenarios (\eg, poster generation). Collectively, our findings reframes few-step distillation from a problem centered primarily on objective design to one that equally demands careful consideration of the training recipe.

\section{Related Work}

\paragraph{Few-step distillation.}
Efficient visual generation is typically achieved through trajectory-level, distribution-level, or hybrid distillation. Trajectory-level methods compress teacher transitions into fewer student steps, including progressive distillation~\citep{salimans2022progressive}, consistency models~\citep{song2023consistency,luo2023latent}, and flow-based methods~\citep{liu2023flow,geng2025mean}. However, pointwise imitation may inherit solver errors and overconstrain the student. Distribution-level methods instead align the student and target distributions, as in adversarial diffusion distillation~\citep{sauer2024adversarial} and DMD~\citep{yin2024improved,yin2024one}, with extensions addressing CFG effects~\citep{liu2025decoupled}, stability~\citep{bai2026optimizing}, and diversity~\citep{wu2026diversity}. TDM~\citep{luo2026tdm} bridges both paradigms by matching distributions along the student trajectory.

\paragraph{Benchmarks for T2I generation and editing.}
T2I models are commonly evaluated using FID and CLIP on MS-COCO~\citep{lin2014microsoft}, as well as fine-grained benchmarks such as GenEval~\citep{ghosh2023geneval} and T2I-CompBench~\citep{huang2023t2icompbench}. However, existing protocols do not fully capture few-step distillation failures in text rendering, composition, prompt adherence, diversity, and visual detail, nor do they jointly evaluate T2I generation and instruction-guided editing. We therefore introduce T2I-Bench and Editing-Bench, two challenging benchmarks that characterize the capabilities and degradation patterns of few-step visual generative models across both tasks.

\section{Preliminaries: Flow Matching and DMD}
We briefly review two components used in this work: flow matching~\citep{lipman2022flow}, a continuous-time framework for learning generative dynamics, and DMD~\citep{yin2024improved}, which we adopt to distill a multi-step teacher into a few-step student.

\subsection{Flow Matching}
Flow matching defines a transport process between data and noise by prescribing a probability path and then learning the velocity field along this path. Let $\bm{x}\sim p_{\text{data}}$ denote a data point and let $\bm{\epsilon}\sim p_{\text{noise}}$ be an independent noise sample, where $p_{\text{noise}}$ is typically set to $\mathcal{N}(\mathbf{0},\mathbf{I})$. In this work, following~\citep{liu2023flow, geng2025mean}, we use the linear path $\bm{z}_t = (1-t)\bm{x} + t\bm{\epsilon}$ for $t\in[0,1]$, which connects the data distribution at $t=0$ to the noise distribution at $t=1$. Under this linear interpolation, the target velocity along each path is $\bm{\epsilon}-\bm{x}$.

For conditional generation, the vector field is
additionally conditioned on $\bm{c}$, which may encode a text prompt and,
in image editing, a source image together with the editing instruction.
Flow matching then trains a parameterized conditional vector field
$\bm{v}_{\bm\theta}(\bm{z}_t,t,\bm{c})$ to recover the target velocity:
\begin{equation}
\ell_{\text{FM}}(\bm\theta) =
\mathbb{E}_{t,\bm{x},\bm{\epsilon}}
\Big[
\big\|
\bm{v}_{\bm\theta}(\bm{z}_t,t,\bm{c})-(\bm{\epsilon}-\bm{x})
\big\|^2
\Big].
\label{eq:fm_loss}
\end{equation}
At inference, samples are generated by drawing $\bm{z}_1\sim p_{\text{noise}}$ and integrating the learned ODE from $t=1$ to $t=0$, yielding $\bm{x}_{\bm\theta}=\bm{z}_1+\int_{1}^{0}\bm{v}_{\bm\theta}(\bm{z}_t,t,\bm{c})\,dt$.

\subsection{DMD Objective}
DMD is designed to distill a pretrained multi-step teacher into a conditional student generator $G_{\bm\theta}$. Given an input noise variable $\bm{\epsilon}$ and condition $\bm c$, the student produces a clean sample
$\bm{x}_{\bm\theta}=G_{\bm\theta}(\bm{\epsilon},\bm c)$. To compare the student with the teacher at noisy intermediate states, an independent noise sample $\bm{\xi}\sim p_{\text{noise}}$ and a noise level $t\sim p_t$ are sampled, where $p_t$ denotes a prescribed distribution over $[0,1]$. The student output is then perturbed as $\bm{x}_t=(1-t)\bm{x}_{\bm\theta}+t\bm{\xi}$.

At a high level, DMD encourages the student-induced conditional distribution to match that of the teacher by minimizing the reverse KL objective $\ell_{\text{DMD}}(\bm{\theta}) \triangleq D_{\mathrm{KL}}\big(p_{\text{stu}}(\bm{x}_{\bm{\theta}}\mid \bm c)\,\|\,p_{\text{tea}}(\bm{x}_{\bm{\theta}}\mid \bm c)\big)$. Rather than optimizing this divergence directly, DMD uses a gradient estimator based on the difference between the score field of the student distribution and that of the teacher distribution:
\begin{equation}
\nabla_{\bm \theta} \ell_{\text{DMD}}(\bm\theta)
=
\mathbb{E}_{\bm{\epsilon},\bm{\xi},t}
\Big[\big(\nabla_{\bm{\theta}} \bm{x}_{\bm{\theta}}\big)^\intercal
\big(
\bm s_{\text{stu}}(\bm{x}_t,t,\bm c)
-
\bm s_{\text{real}}(\bm{x}_t,t,\bm c)
\big)
\Big].
\label{eq:dmd}
\end{equation}
Here, $\bm s_{\text{stu}}$ is estimated using an auxiliary score network trained on samples generated by the student, while $\bm s_{\text{real}}$ is obtained from the pretrained teacher. The resulting update pushes the student toward regions where its noisy marginal score agrees with the teacher score across sampled noise levels.

\section{Experiments}
This section introduces T2I-Bench and Editing-Bench for few-step generation and editing, then uses Qwen-Image-2.0~\citep{zhao2026qwen} as a testbed to systematically study how training data composition, teacher guidance, and task mixture shape DMD, culminating in \textbf{Qwen-Image-Flash}.

\subsection{Benchmarks and Evaluation Protocol}

\paragraph{T2I-Bench.}
T2I-Bench is a challenging evaluation suite covering the same three categories considered in our data-composition study. It comprises $1{,}800$ evaluation cases, evenly distributed across categories with $600$ cases each. We employ Gemini 3.1 Pro and GPT 5.5 as automatic preference-based evaluators, where higher scores indicate superior perceptual quality, prompt alignment, and overall alignment with human preferences. The rubric-based evaluation prompt and additional details are provided in the Appendix.

\paragraph{Editing-Bench.}
Editing-Bench comprises $1{,}500$ challenging cases evenly distributed across six representative categories: scene-level semantic transformation, perceptual enhancement, object-centric manipulation, textual content editing, identity-preserving editing, and stylistic transfer. With $250$ cases per category, it supports balanced evaluation of global and local editing capabilities. Following T2I-Bench, we employ Gemini 3.1 Pro and GPT 5.5 as automatic evaluators, with higher scores indicating better instruction following, source-image preservation, artifact suppression, and alignment with human preferences. Category-specific system prompts capture distinct task-level quality criteria. Further details appear in the Appendix.

\subsection{Data Composition Matters in T2I Distillation}
\paragraph{Experimental setup.}
\label{sec:data_composition}
We use Qwen-Image-2.0-Base~\citep{zhao2026qwen} as the multi-step teacher and distill it into a $4$-NFE student with DMD~\citep{yin2024improved}. To construct the training data for distillation, we use Qwen3~\citep{yang2025qwen3} to generate $20{,}000$ diverse prompts for each of three representative categories: landscapes, portraits, and text-centric scenarios. From these category-specific training sets, we derive five data compositions with varying category coverage: landscape-only, portrait-only, text-centric-only, landscape-portrait, and a mixture of all three categories. All students are trained under the same optimization protocol for $2{,}000$ iterations using AdamW~\citep{loshchilov2017decoupled}, with a learning rate of $1\times 10^{-6}$ and a batch size of $64$ for both the student and the auxiliary score network, so that performance differences can be attributed primarily to the training data composition.

\begin{table*}[t]
\centering
\caption{\textbf{T2I distillation with different data compositions.} We evaluate $4$-NFE students distilled with different category-specific and mixed-category training sets on landscape, portrait, and text-centric splits of T2I-Bench.}
\label{tab:t2i_data_composition}
\tablestyle{3.5pt}{1.2}
\begin{tabular}{c|c|c|ccc|c|c}
\multirow{2}{*}{\begin{tabular}[c]{@{}c@{}}\textbf{Training data}\\\textbf{composition}\end{tabular}}
& \multirow{2}{*}{\begin{tabular}[c]{@{}c@{}}\textbf{\# of}\\\textbf{training data}\end{tabular}}
& \multirow{2}{*}{\textbf{Metrics}}
& \multicolumn{3}{c|}{\textbf{T2I-Bench}}
& \multirow{2}{*}{\textbf{Avg}}
& \multirow{2}{*}{\textbf{Rank}} \\
& & & \textbf{Landscape} & \textbf{Portrait} & \textbf{Text-centric} & & \\ \shline

\multirow{2}{*}{Landscape}
& \multirow{2}{*}{20,000}
& Gemini 3.1 Pro
& 3.53
& 3.37
& 3.01
& 3.30
& \multirow{2}{*}{3} \\
&
& GPT 5.5
& 4.30
& 4.31
& 3.77
& 4.13
& \\ \midline

\multirow{2}{*}{Portrait}
& \multirow{2}{*}{20,000}
& Gemini 3.1 Pro
& 3.56
& 3.57
& 3.12
& 3.42
& \multirow{2}{*}{1} \\
&
& GPT 5.5
& 4.35
& 4.34
& 3.76
& 4.15
& \\ \midline

\multirow{2}{*}{Text-centric}
& \multirow{2}{*}{20,000}
& Gemini 3.1 Pro
& 2.55
& 3.38
& 1.97
& 2.63
& \multirow{2}{*}{5} \\
&
& GPT 5.5
& 3.34
& 3.88
& 2.64
& 3.29
& \\ \midline

\multirow{2}{*}{Landscape-portrait}
& \multirow{2}{*}{40,000}
& Gemini 3.1 Pro
& 3.61
& 3.54
& 3.04
& 3.40
& \multirow{2}{*}{2} \\
&
& GPT 5.5
& 4.24
& 4.33
& 3.62
& 4.06
& \\ \midline

\multirow{2}{*}{Mixed-category}
& \multirow{2}{*}{60,000}
& Gemini 3.1 Pro
& 3.53
& 3.47
& 2.05
& 3.02
& \multirow{2}{*}{4} \\
&
& GPT 5.5
& 4.08
& 4.23
& 2.54
& 3.62
&
\end{tabular}
\end{table*}

\begin{figure*}[t]
\begin{center}
\vspace{-1mm}
\includegraphics[width=\linewidth]{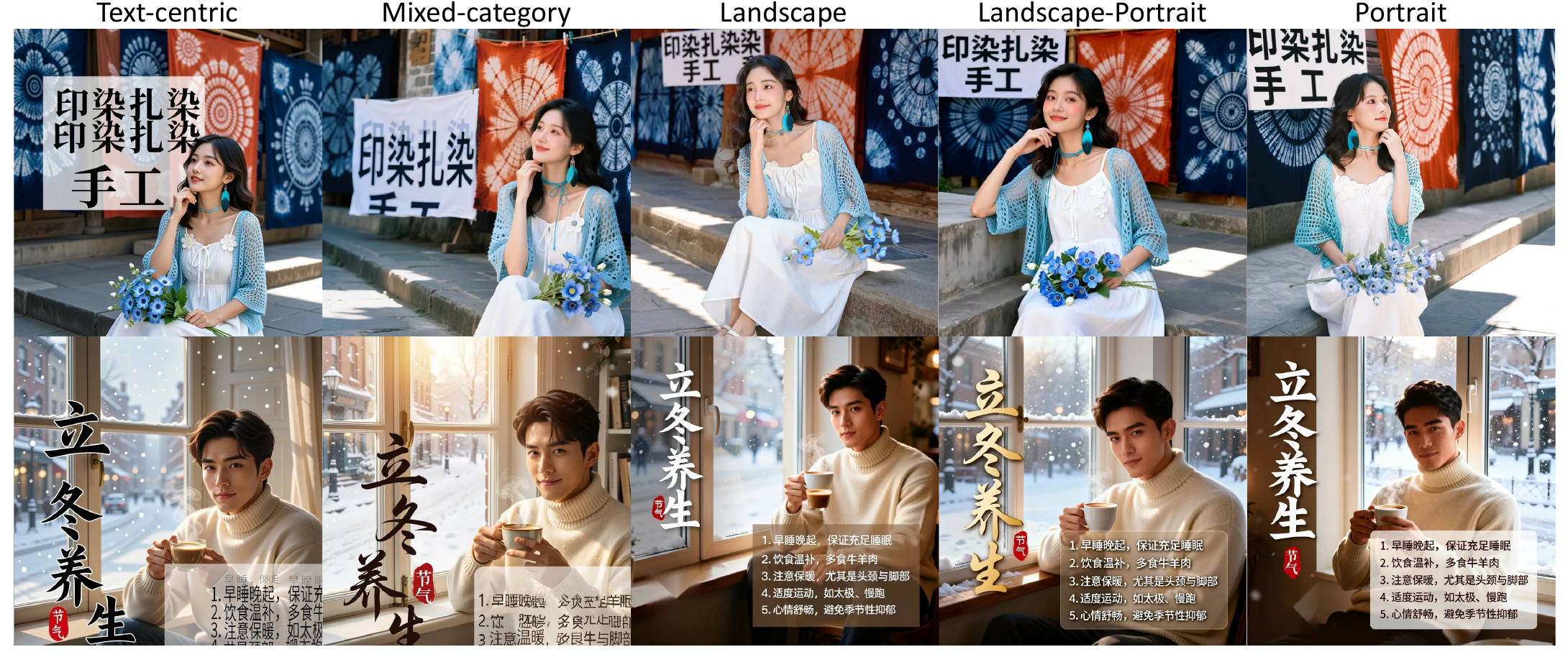}
\caption{\textbf{Qualitative results under different data compositions.} We compare students distilled with text-centric, mixed-category, landscape-only, landscape-portrait, and portrait-only training data across representative evaluation scenarios.}
\label{fig:training_data}
\end{center}
\end{figure*}

\paragraph{Target-aligned data can be unexpectedly detrimental.} Table~\ref{tab:t2i_data_composition} shows that the text-centric setting provides the most direct exposure to text-heavy samples yet yields the lowest overall performance. Surprisingly, this degradation extends in-domain: on the text-centric split, it is substantially outperformed by both landscape- and portrait-based distillation, neither of which uses explicit text-centric training data. The mixed-category setting exhibits a similar trend: despite using the largest training set and covering all three categories, it underperforms the stronger single-category configurations, particularly on text-centric evaluation. These results challenge the \textit{common intuition} from large-scale pretraining that better target coverage and greater data diversity should consistently improve performance. In few-step distillation, certain data distributions may instead hinder distribution transfer, as also illustrated qualitatively in Figure~\ref{fig:training_data}.

\begin{tcolorbox}[
colback=blue!3,
colframe=blue!70!black,
boxrule=0.8pt,
arc=6pt,
left=6pt,
right=6pt,
top=4pt,
bottom=4pt
]
\textbf{Takeaway 1:} Unlike the common scaling intuition in pretraining, broader or target-aligned data does not necessarily benefit few-step distillation and may even hurt student performance.
\end{tcolorbox}

\paragraph{Category-specific data can support broad transfer.}
In contrast, single-category distillation exhibits strong cross-category generalization. Landscape- and portrait-based students remain competitive beyond their training domains and even outperform text-centric and mixed-category settings on text-centric evaluation without seeing such prompts during training. Moreover, combining landscape and portrait data provides no further gain over portrait-only distillation. These results suggest that explicit downstream coverage is not necessary for effective few-step distillation, and that broader category coverage does not necessarily translate into better transfer.

\begin{tcolorbox}[
colback=blue!3,
colframe=blue!70!black,
boxrule=0.8pt,
arc=6pt,
left=6pt,
right=6pt,
top=4pt,
bottom=4pt
]
\textbf{Takeaway 2:} Data selection is critical in few-step T2I distillation: single-category data can generalize broadly, while broader category combinations do not guarantee further gains.
\end{tcolorbox}

\subsection{Transferring Post-trained Capabilities through Teacher Guidance}
\label{sec:step_wise_guidance}
Beyond training data composition, effective distillation also depends on how teacher supervision is incorporated. As post-training becomes increasingly common for enhancing visual generative models, a practical question arises: whether and how these improvements can be retained through few-step distillation. We therefore consider a post-trained multi-step teacher and ask: \textit{how should its supervision be used to reliably transfer these gains to a few-step student?} This motivates the second dimension of our DMD~\citep{yin2024improved} training recipe: \textit{teacher guidance}.

\paragraph{Experimental setup.}
Following the post-training paradigm of \citet{xu2026qwen}, we further post-train Qwen-Image-2.0-Base~\citep{zhao2026qwen} to obtain a task-specialized teacher, termed Qwen-Image-2.0-RL, with enhanced downstream capabilities. We adopt the same experimental protocol as in Section~\ref{sec:data_composition}, using the portrait-only training set and identical optimization hyperparameters.

\begin{figure*}[t]
\begin{center}
\includegraphics[width=\linewidth]{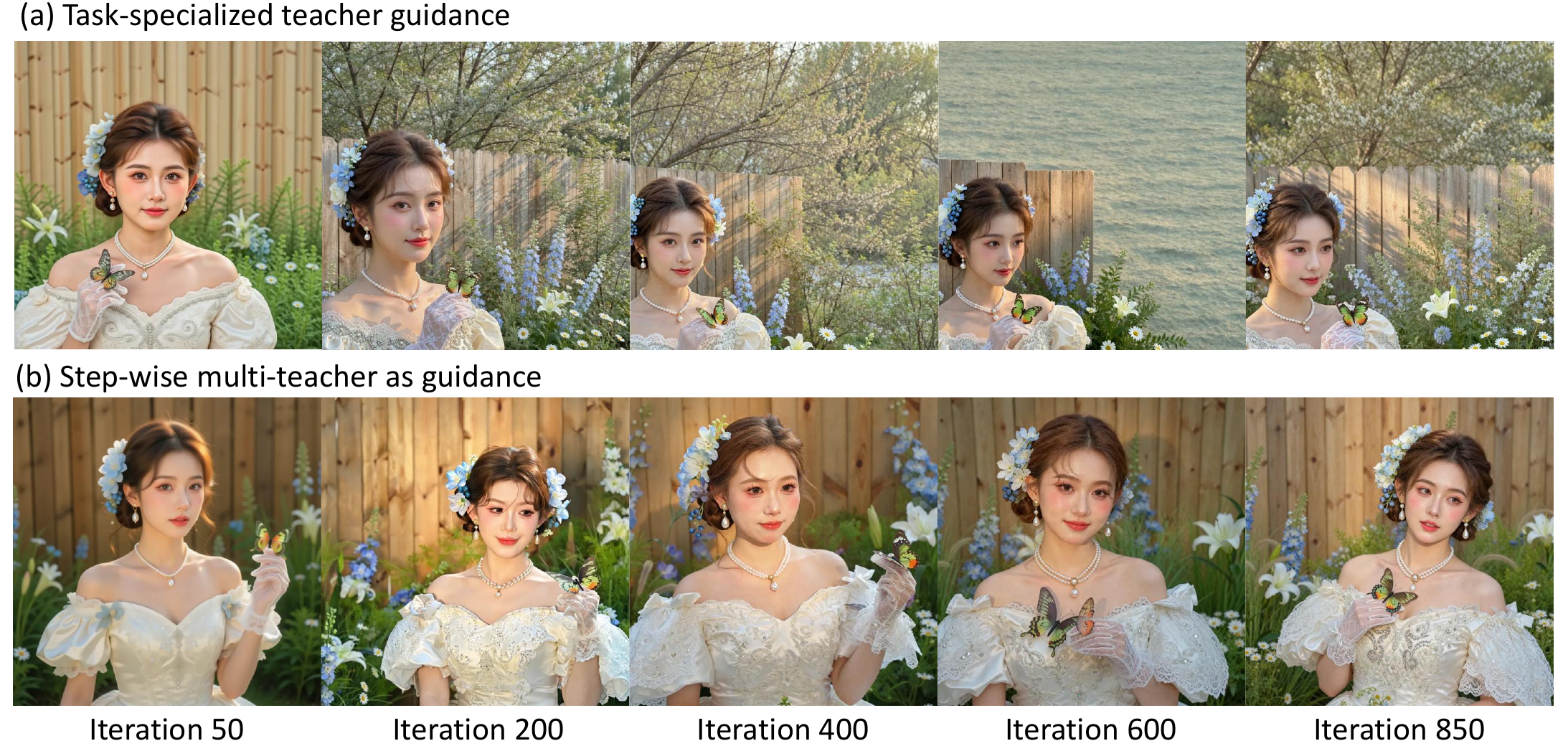}
\caption{\textbf{Qualitative comparison of teacher guidance strategies during distillation.} (a) Direct guidance from a task-specialized teacher can destabilize training, leading to progressive degradation in alignment and visual quality. (b) Step-wise multi-teacher guidance maintains sample fidelity and layout consistency throughout distillation, yielding better-aligned generations.}
\label{fig:multi-teacher_observation}
\end{center}
\end{figure*}

\paragraph{Direct post-trained teacher guidance destabilizes distillation.}
\label{sec:mtg_observation}
As shown in Figure~\ref{fig:multi-teacher_observation} (a), directly distilling from the task-specialized teacher results in pronounced optimization instability. While the student initially exhibits task-specific gains, generation quality progressively degrades during training, manifesting as structural distortion, reduced visual fidelity, and weaker semantic alignment. This observation indicates that stronger downstream teacher performance does not necessarily translate into more effective guidance for few-step distillation. We conjecture that post-training induces a distribution that is more difficult for a highly compressed student to approximate, thereby amplifying the mismatch between teacher and student score fields under DMD and ultimately leading to unstable optimization or collapse.

\begin{tcolorbox}[
colback=blue!3,
colframe=blue!70!black,
boxrule=0.8pt,
arc=6pt,
left=6pt,
right=6pt,
top=4pt,
bottom=4pt
]
\textbf{Takeaway 3:} Directly using a post-trained teacher as the sole source of guidance can destabilize few-step distillation, despite its stronger downstream performance.
\end{tcolorbox}

\paragraph{Step-wise multi-teacher guidance.}
Motivated by this observation, we retain the pretrained base teacher as a stable distributional anchor within DMD and selectively incorporate post-trained, task-specialized guidance across student distillation steps. This recipe draws on recent on-policy distillation methods~\citep{xiao2026mimo, li2026diffusionopd, fang2026flow}, which emphasize adapting supervision to the evolving student policy. 

More generally, let $T_0$ denote the base teacher and $\{T_m\}_{m=1}^{M}$ denote task-specialized teachers. At the $k$-th selected student distillation step, we construct the DMD real-score target as
$\bm{s}_{\mathrm{real}}^{(k)}(\bm{x}_t,t,\bm{c})
=
\sum_{m=0}^{M}
\lambda_{k,m}(\bm{c})\,
\bm{s}_m^{(k)}(\bm{x}_t,t,\bm{c})$,
where $\lambda_{k,m}(\bm{c}) \in [0,1]$ controls the contribution of teacher $T_m$ and satisfies
$\sum_{m=0}^{M}\lambda_{k,m}(\bm{c})=1$.
In this work, we consider the simplest non-trivial case, $M=1$, with one pretrained base teacher and one post-trained task-specialized teacher. Step-dependent weighting preserves stable base supervision while selectively introducing specialized capabilities along the student trajectory. The resulting optimization at step $k$ is formulated as
\begin{equation}
\nabla_{\bm \theta} \ell_{\text{DMD}}^{(k)}(\bm\theta)
=
\mathbb{E}_{\bm{\epsilon},\bm{\xi},t}
\Big[
\big(\nabla_{\bm{\theta}} \bm{x}_{\bm{\theta}}\big)^\intercal
\big(
\bm s_{\text{stu}}(\bm{x}_t,t,\bm c)
-
\sum_{m=0}^{M}
\lambda_{k,m}(\bm c)
\bm s_m^{(k)}(\bm{x}_t,t,\bm c)
\big)
\Big].
\label{eq:modified_dmd}
\end{equation}

\begin{table*}[t]
\centering
\caption{\textbf{Quantitative comparison of teachers and student.} With only $4$ NFEs, the distilled model achieves competitive performance against 80-NFE teachers, effectively inheriting their complementary strengths across landscape, portrait, and text-centric evaluation sets.}
\label{tab:t2i_multiteacher}
\tablestyle{3.5pt}{1.2}
\begin{tabular}{c|c|c|ccc|c|c}
\multirow{2}{*}{\textbf{Model}}      
& \multirow{2}{*}{\textbf{NFEs}} 
& \multirow{2}{*}{\textbf{Metrics}} 
& \multicolumn{3}{c|}{\textbf{T2I-Bench}}        
& \multirow{2}{*}{\textbf{Avg}} 
& \multirow{2}{*}{\textbf{Rank}} \\
& & & \textbf{Landscape} & \textbf{Portrait} & \textbf{Text-centric} & & \\ \shline

\multirow{2}{*}{Qwen-Image-2.0-Base} 
& \multirow{2}{*}{80}            
& Gemini 3.1 Pro    
& 3.52  
& 3.64  
& 3.08   
& 3.41
& \multirow{2}{*}{3} \\
& 
& GPT 5.5           
& 4.24  
& 4.30 
& 3.73    
& 4.09
& \\ \midline

\multirow{2}{*}{Qwen-Image-2.0-RL}   
& \multirow{2}{*}{80}            
& Gemini 3.1 Pro    
& 3.98   
& 3.82  
& 3.41   
& 3.74
& \multirow{2}{*}{1} \\
& 
& GPT 5.5           
& 4.34   
& 4.47 
& 3.88    
& 4.26
& \\ \midline

\multirow{2}{*}{\textbf{Qwen-Image-Flash-T2I}}
& \multirow{2}{*}{4}             
& Gemini 3.1 Pro    
& 3.88   
& 3.81  
& 3.00     
& 3.56
& \multirow{2}{*}{2} \\
& 
& GPT 5.5           
& 4.30  
& 4.41 
& 3.75    
& 4.15
& \\
\end{tabular}
\end{table*}

\paragraph{Training hyperparameters.}
We instantiate step-wise multi-teacher guidance with a $4$-step student, \ie, $k\in\{1,2,3,4\}$. To mitigate the severe structural degradation observed in Figure~\ref{fig:multi-teacher_observation} (a) under direct post-trained teacher guidance, the first three steps use only the pretrained base teacher for stable structural supervision: $\lambda_{k,0}=1$ and $\lambda_{k,1}=0$ for $k\in\{1,2,3\}$. At the final step, once a stable global structure is established, we equally combine both teachers, setting $\lambda_{4,0}=\lambda_{4,1}=0.5$. This schedule preserves the base teacher's structural stability while selectively transferring the post-trained teacher's enhanced downstream capabilities, yielding \textbf{Qwen-Image-Flash-T2I}.

\paragraph{Main results.}
Figure~\ref{fig:multi-teacher_observation} (b) shows that step-wise multi-teacher guidance effectively stabilizes student optimization, maintaining sample fidelity, layout consistency, and semantic alignment throughout training, in contrast to the progressive degradation observed with direct specialized-teacher distillation. Table~\ref{tab:t2i_multiteacher} further demonstrates effective capability transfer: with only $4$ NFEs, \textbf{Qwen-Image-Flash-T2I} achieves average scores of $3.56$ with Gemini 3.1 Pro and $4.15$ with GPT 5.5, surpassing the $80$-NFE Qwen-Image-2.0-Base teacher in the overall ranking. These results suggest that anchoring early distillation with the base teacher while selectively introducing specialized supervision provides an effective balance between structural preservation and downstream capability transfer. The strategy is also \textit{lightweight and general}, requiring only step-dependent guidance weights to incorporate specialized teachers into existing DMD pipelines.

\begin{tcolorbox}[
colback=blue!3,
colframe=blue!70!black,
boxrule=0.8pt,
arc=6pt,
left=6pt,
right=6pt,
top=4pt,
bottom=4pt
]
\textbf{Takeaway 4:} Incorporating post-trained teachers is a scheduling problem rather than a replacement problem: anchor early distillation with the base teacher and introduce specialized guidance only at later stages.
\end{tcolorbox}

\subsection{Joint Distillation for Unified Generation and Editing}
Beyond T2I generation, the teacher model also supports instruction-guided image editing through the same backbone~\citep{zhao2026qwen}, making both tasks integral to the capabilities we seek to preserve through distillation. We therefore jointly distill generation and editing into a single few-step student to preserve the teacher's unified capabilities. As the tasks provide distinct training signals, overemphasizing either may compromise the other, making task mixture critical. We therefore investigate the T2I-to-editing data ratio to balance performance across both tasks.

\paragraph{Experimental setup.}
We perform joint distillation with three T2I-to-editing data ratios: $9{:}1$, $7{:}3$, and $5{:}5$, ranging from T2I-dominant to balanced mixtures, using an internal dataset for editing supervision. All other training hyperparameters and settings follow Section~\ref{sec:step_wise_guidance}, with the total training budget held constant and only the task ratio varied.

\begin{table*}[t]
\centering
\caption{\textbf{Effect of T2I:Edit ratios on Editing-Bench across six editing dimensions.} Tea. denotes the multi-step teacher trained on the same internal T2I and editing datasets and distilled for faster inference. Zero-shot denotes Qwen-Image-Flash-T2I without editing distillation.}
\label{tab:editing_bench_result}
\tablestyle{3.5pt}{1.2}
\resizebox{0.99\textwidth}{!}{%
\begin{tabular}{c|c|cccccc|c|c}
\multirow{2}{*}{\textbf{Ratio}}
& \multirow{2}{*}{\textbf{Metrics}} 
& \multicolumn{6}{c|}{\textbf{Editing-Bench}}       
& \multirow{2}{*}{\textbf{Avg}}
& \multirow{2}{*}{\textbf{Rank}} \\
& & \begin{tabular}[c]{@{}c@{}}\textbf{Scene}\\\textbf{trans.}\end{tabular}
& \begin{tabular}[c]{@{}c@{}}\textbf{Perceptual}\\\textbf{enhance.}\end{tabular}
& \begin{tabular}[c]{@{}c@{}}\textbf{Object}\\\textbf{manip.}\end{tabular}
& \begin{tabular}[c]{@{}c@{}}\textbf{Text}\\\textbf{editing}\end{tabular}
& \begin{tabular}[c]{@{}c@{}}\textbf{Identity}\\\textbf{preserv.}\end{tabular}
& \begin{tabular}[c]{@{}c@{}}\textbf{Style}\\\textbf{transfer}\end{tabular} & & \\ \shline

\multirow{2}{*}{Tea.} 
& Gemini 3.1 Pro     
& 2.52 & 2.75 & 2.33 & 3.25 & 3.16 & 2.62 & 2.77 & \multirow{2}{*}{3} \\
& GPT 5.5 
& 3.61 & 3.22 & 3.05 & 3.69 & 3.53 & 3.56 & 3.44 & \\ \midline

\multirow{2}{*}{Zero-shot} 
& Gemini 3.1 Pro     
& 2.78 & 2.81 & 2.25 & 2.84 & 3.16 & 2.78 & 2.77 & \multirow{2}{*}{4} \\
& GPT 5.5 
& 3.57 & 3.03 & 3.01 & 3.11 & 3.41 & 3.52 & 3.28 & \\ \midline

\multirow{2}{*}{9:1} 
& Gemini 3.1 Pro     
& 2.43 & 2.25 & 2.09 & 2.83 & 3.12 & 2.75 & 2.58 & \multirow{2}{*}{5} \\
& GPT 5.5 
& 3.48 & 3.21 & 2.87 & 3.25 & 3.53 & 3.50 & 3.31 & \\ \midline

\multirow{2}{*}{7:3}   
& Gemini 3.1 Pro      
& 2.49 & 2.80 & 2.34 & 3.35 & 3.18 & 3.05 & 2.87 & \multirow{2}{*}{2} \\
& GPT 5.5 
& 3.46 & 2.87 & 3.08 & 3.58 & 3.52 & 3.63 & 3.36 & \\ \midline

\multirow{2}{*}{5:5} 
& Gemini 3.1 Pro     
& 2.86 & 2.25 & 2.68 & 3.18 & 3.19 & 3.68 & 2.97 & \multirow{2}{*}{1} \\
& GPT 5.5 
& 3.66 & 3.06 & 3.20 & 3.13 & 3.47 & 3.92 & 3.41 & \\ 

\end{tabular}%
}
\end{table*}

\begin{figure*}[t]
\begin{center}
\includegraphics[width=\linewidth]{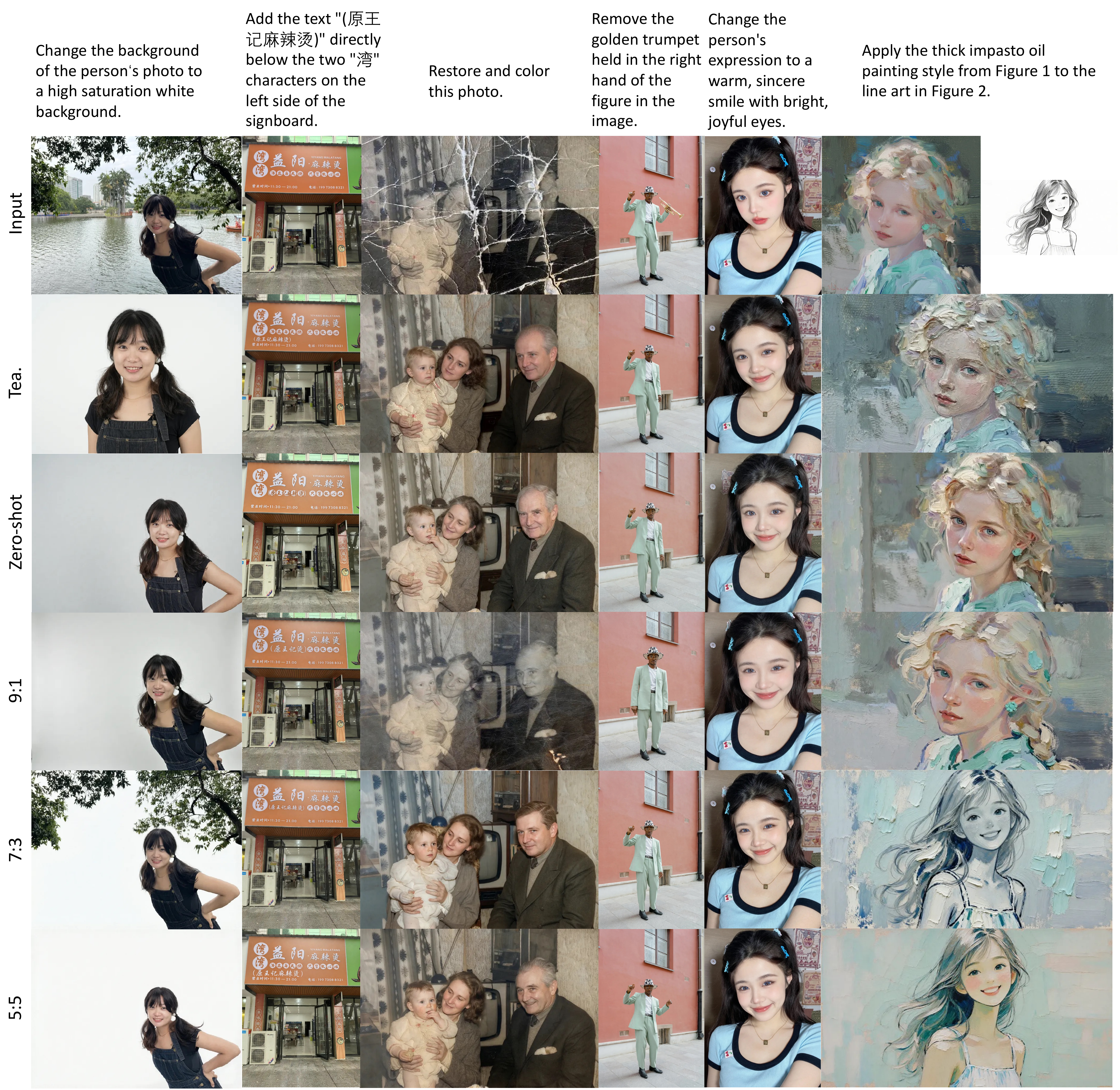}
\caption{\textbf{Qualitative comparison under different T2I:Edit ratios.} Across six editing categories, we compare the editing teacher, T2I-only zero-shot student, and jointly distilled students trained with $9{:}1$, $7{:}3$, and $5{:}5$ mixtures. The balanced $5{:}5$ ratio best improves instruction following while preserving fidelity, identity, and style.}
\label{fig:editing_ratio}
\end{center}
\end{figure*}

\begin{table*}[t]
\centering
\caption{\textbf{Effect of T2I:Edit ratios on T2I-Bench.} We evaluate how editing supervision affects T2I generation in jointly distilled students.
}
\label{tab:joint_t2i_bench_result}
\tablestyle{3.5pt}{1.2}
\begin{tabular}{c|c|cccccc|c|c}
\multirow{2}{*}{\textbf{Ratio}}
& \multirow{2}{*}{\textbf{Metrics}} 
& \multicolumn{6}{c|}{\textbf{T2I-Bench}}       
& \multirow{2}{*}{\textbf{Avg}}
& \multirow{2}{*}{\textbf{Rank}} \\
& & \multicolumn{2}{c}{\textbf{Landscape}} & \multicolumn{2}{c}{\textbf{Portrait}} & \multicolumn{2}{c|}{\textbf{Text-centric}} & & \\ \shline

\multirow{2}{*}{Qwen-Image-2.0-RL} 
& Gemini 3.1 Pro     
& \multicolumn{2}{c}{3.98}   
& \multicolumn{2}{c}{3.82}  
& \multicolumn{2}{c|}{3.41}       
& 3.74
& \multirow{2}{*}{1} \\
& GPT 5.5 
& \multicolumn{2}{c}{4.34} 
& \multicolumn{2}{c}{4.47}  
& \multicolumn{2}{c|}{3.88}   
& 4.26
& \\ \midline

\multirow{2}{*}{Qwen-Image-Flash-T2I} 
& Gemini 3.1 Pro     
& \multicolumn{2}{c}{3.88}   
& \multicolumn{2}{c}{3.81}  
& \multicolumn{2}{c|}{3.00}       
& 3.56
& \multirow{2}{*}{5} \\
& GPT 5.5 
& \multicolumn{2}{c}{4.30} 
& \multicolumn{2}{c}{4.41}  
& \multicolumn{2}{c|}{3.75}   
& 4.15
& \\ \midline

\multirow{2}{*}{Qwen-Image-Flash (9:1)} 
& Gemini 3.1 Pro     
& \multicolumn{2}{c}{4.04}   
& \multicolumn{2}{c}{3.58}  
& \multicolumn{2}{c|}{3.17} 
& 3.60
& \multirow{2}{*}{4} \\
& GPT 5.5 
& \multicolumn{2}{c}{4.39}   
& \multicolumn{2}{c}{4.37}  
& \multicolumn{2}{c|}{3.77} 
& 4.18
& \\ \midline

\multirow{2}{*}{Qwen-Image-Flash (7:3)}   
& Gemini 3.1 Pro      
& \multicolumn{2}{c}{3.81}   
& \multicolumn{2}{c}{3.77}  
& \multicolumn{2}{c|}{3.21}   
& 3.60
& \multirow{2}{*}{3} \\
& GPT 5.5 
& \multicolumn{2}{c}{4.35}   
& \multicolumn{2}{c}{4.40}  
& \multicolumn{2}{c|}{3.84}   
& 4.20
& \\ \midline

\multirow{2}{*}{\textbf{Qwen-Image-Flash} (5:5)} 
& Gemini 3.1 Pro     
& \multicolumn{2}{c}{3.95}   
& \multicolumn{2}{c}{3.85}  
& \multicolumn{2}{c|}{3.14}       
& 3.65
& \multirow{2}{*}{2} \\
& GPT 5.5 
& \multicolumn{2}{c}{4.34} 
& \multicolumn{2}{c}{4.37}  
& \multicolumn{2}{c|}{3.76}   
& 4.16
& \\ 

\end{tabular}
\end{table*}

\paragraph{Balanced mixtures improve editing transfer.}
As shown in Table~\ref{tab:editing_bench_result}, the T2I-only student retains non-trivial zero-shot editing ability, but still underperforms the specialized teacher, particularly on text editing. Adding limited editing data is insufficient: the $9{:}1$ mixture performs worse than the T2I-only baseline. Performance improves as editing supervision increases, with $7{:}3$ surpassing both the baseline and teacher under Gemini 3.1 Pro, and $5{:}5$ achieving the best overall results. Specifically, the balanced mixture improves the average score from $2.77$ to $2.97$ under Gemini 3.1 Pro and from $3.28$ to $3.41$ under GPT 5.5. Figure~\ref{fig:editing_ratio} further confirms its stronger instruction following and source preservation. We therefore adopt the $5{:}5$ task ratio for our final unified model, \textbf{Qwen-Image-Flash}.

\begin{tcolorbox}[
colback=blue!3,
colframe=blue!70!black,
boxrule=0.8pt,
arc=6pt,
left=6pt,
right=6pt,
top=4pt,
bottom=4pt
]
\textbf{Takeaway 5:} Editing transfer is highly sensitive to the task ratio: limited editing data can be ineffective, whereas a balanced mixture yields the strongest editing performance.
\end{tcolorbox}

\paragraph{Editing supervision improves T2I generation.}
We also evaluate the T2I performance of the jointly distilled students to examine how introducing editing data affects generation quality. Table~\ref{tab:joint_t2i_bench_result} shows that all jointly distilled students outperform the T2I-only baseline on average T2I-Bench scores. This positive transfer suggests that editing data provides complementary supervision for instruction grounding, spatial localization, and visual-textual consistency, thereby strengthening prompt following and semantic alignment in T2I generation.

\begin{tcolorbox}[
colback=blue!3,
colframe=blue!70!black,
boxrule=0.8pt,
arc=6pt,
left=6pt,
right=6pt,
top=4pt,
bottom=4pt
]
\textbf{Takeaway 6:} Editing supervision provides positive cross-task transfer, improving T2I generation rather than merely preserving it.
\end{tcolorbox}

\section{Discussion and Conclusion}
\paragraph{Limitations and future work.}
Despite competitive T2I generation and instruction-guided editing with only $4$ NFEs, Qwen-Image-Flash exhibits minor text-rendering artifacts and residual noise. Our study is primarily constrained by its empirical scope, covering a single model family, a fixed sampling budget, and limited teacher schedules. Inspired by the step-specific roles in DP-DMD~\citep{wu2026diversity}, future work will explore combining trajectory matching for structural preservation with distribution matching for fine-detail refinement to improve training stability and visual fidelity under aggressive sampling compression.

\paragraph{Conclusion.}
In this work, we systematically revisit the training recipe for few-step distillation of visual generative foundation models, using Qwen-Image-2.0 as an empirical testbed within the DMD framework. Our study identifies \textit{data composition, teacher guidance, and task mixture} as critical determinants of effective capability transfer: broader or target-aligned data can hinder distillation, step-wise multi-teacher guidance reconciles structural stability with post-trained capabilities, and balanced task supervision preserves unified generation and editing. These insights culminate in \textbf{Qwen-Image-Flash}, which retains strong T2I generation and instruction-guided editing performance with only $4$ NFEs. Collectively, our findings broaden the perspective on few-step distillation by demonstrating how the training recipe shapes the realization of a distillation objective. As visual foundation models grow in capability and scope, principled \textbf{training recipe} design will be essential to translating their advances into efficient, stable, and broadly capable generative systems.

\section*{AI Use Statement}
In this work, we used generative AI tools to assist with research code implementation. We have not used generative AI tools to develop conceptual frameworks, formulate hypotheses, design experiments, or interpret results; tasks related to theoretical proofs are not applicable to this work. Additionally, we used generative AI tools to polish the manuscript for clarity, readability, and linguistic consistency. We have carefully reviewed all AI-assisted work, including manually inspecting the generated code for correctness and reviewing text revisions for accuracy and consistency with our intended scientific meaning. We take full responsibility for the final content of this work, including all text, claims, and artifacts produced with the assistance of generative AI.

\section*{Ethics Statement}
This work advances inference acceleration for visual generative models, improving the efficiency and accessibility of image generation and editing. Our methods are intended for legitimate research and beneficial applications, not harmful or deceptive uses. We advocate responsible use that respects privacy, intellectual property, and ethical standards.

\section*{Reproducibility Statement}
The main paper describes the distillation objective, training data compositions, teacher guidance schedule, task mixtures, and optimization settings used in our experiments. Benchmark descriptions and evaluation protocols are provided in the main text, with detailed evaluator prompts and category-specific rubrics in the Appendix. These details support reproducibility, although full replication requires access to the underlying teacher models and internal editing data.


\bibliography{iclr2027_conference}
\bibliographystyle{iclr2027_conference}

\clearpage
\appendix
\section*{Appendix}

\renewcommand{\thefigure}{\Alph{figure}}
\renewcommand{\thetable}{\Alph{table}}
\setcounter{figure}{0}
\setcounter{table}{0}

\section{Benchmark Evaluation Protocols}
\label{app:evaluation_details}
This section details the evaluation protocols for T2I-Bench and Editing-Bench, including evaluator prompts, scoring rubrics, and structured output formats.

\subsection{Evaluator Prompts and Scoring Rubrics}
\label{app:system_prompts}
We employ Gemini 3.1 Pro and GPT 5.5 as automatic evaluators, guided by benchmark-specific instructions that specify the assessment criteria and scoring principles. The full prompts and task-specific rubrics are provided below.

\paragraph{T2I-Bench evaluation.}
Given a text prompt and its generated image, each evaluator follows the system prompt in Table~\ref{tab:text_prompt} to assess two equally important dimensions: \emph{prompt compliance} and \emph{technical quality}. The former examines whether the image correctly depicts the specified objects, attributes, quantities, actions, and spatial relationships. The latter assesses structural and perceptual defects, including geometric inconsistencies, anatomical errors, corrupted text, and texture degradation. Aesthetic appeal should not compensate for failures to satisfy the requested content.

Table~\ref{tab:t2i_bench_hard_cases} provides two challenging examples involving detailed portrait specifications and text-rich structured compositions. Each assessment is returned as strict JSON containing an overall score from $1$ to $5$, rounded to two decimal places, and a concise, single-sentence justification.

\paragraph{Editing-Bench evaluation.}
For Editing-Bench, each evaluator receives the source image, editing instruction, and edited output. We use a shared system prompt template with category-specific rubrics to assess both the requested modification and the preservation of source content that should remain unchanged.

The template in Table~\ref{tab:image_editing_prompt_template} specifies the common scoring scale, output schema, and scoring principles. For each sample, it is instantiated according to the task category through three placeholders: \texttt{\textless{}category-title\textgreater{}}, \texttt{\textless{}category-rubric\textgreater{}}, and \texttt{\textless{}sub-score-criteria\textgreater{}}. Table~\ref{tab:image_editing_prompt_placeholders} defines the category-specific assessment priorities. For example, perceptual enhancement emphasizes faithful detail recovery without hallucinated content, whereas object-centric manipulation emphasizes accurate localization and natural integration.

Table~\ref{tab:image_editing_prompt_subscore_criteria} lists the corresponding criteria for individual sub-scores, providing diagnostic detail beyond the overall assessment. The evaluator returns strict JSON containing an overall score from $1$ to $5$, rounded to two decimal places, criterion-level sub-scores, a single-sentence justification, and identified failure modes. The overall score reflects a holistic assessment rather than an arithmetic average of sub-scores, with severe failures in core requirements, such as instruction following, identity preservation, or text accuracy, strongly limiting the final score.

\begin{table}[b]
\centering
\renewcommand{\arraystretch}{1}
\caption{\textbf{T2I-Bench evaluation.} System prompt and scoring criteria.}
\label{tab:text_prompt}
\resizebox{1.0\textwidth}{!}{
\begin{tabular}{p{\textwidth}}
\toprule
You are an expert evaluator for text-to-image generation results. Rate the image based on two equally critical dimensions: instruction following and visible technical quality. Do not reward an image for being merely beautiful if it fails to depict the prompt accurately. \\
\\
\textbf{Prompt Compliance}: Verify if the image precisely reflects all objects, attributes, quantities, actions, and spatial relationships described in the caption. \\
\textbf{Technical Quality}: Focus on structural and perceptual defects: distorted geometry, warped or broken objects, inconsistent perspective, bad anatomy, face, hand, or limb deformation, texture melting, duplicated or missing parts, corrupted text or logos, blur, noise, compression artifacts, aliasing, exposure problems, color casts, unrealistic lighting, and other image-generation artifacts. \\
\\
Return strict JSON only, with no markdown: \{"score": \textless{}float from 1 to 5 rounded to two decimals\textgreater{}, "reasoning": "\textless{}one short English sentence\textgreater{}"\}. \\
\bottomrule
\end{tabular}
}
\end{table}

\begin{table}[H]
\centering
\renewcommand{\arraystretch}{1.12}
\caption{\textbf{Challenging prompts from T2I-Bench.} Randomly selected examples illustrating complex generation requirements across diverse scenarios.}
\label{tab:t2i_bench_hard_cases}
\resizebox{0.9\textwidth}{!}{
\begin{tabular}{p{0.12\textwidth}p{0.86\textwidth}}
\toprule
Sample & Prompt \\
\midrule
\texttt{1} &
A young East Asian woman, about 20 to 28 years old, sits gracefully and casually on outdoor cement steps. She has medium-length wavy black hair with slightly curled ends falling naturally over both shoulders, a soft oval face, fair smooth skin without obvious moles or freckles, and clean natural makeup: light brown eyeshadow, thin inner eyeliner, thick curled lashes, soft pink-orange blush, matte coral-pink lips, a slight smile, and a calm cheerful gaze looking toward the upper right. She wears a white cotton camisole dress with two three-dimensional white fabric flowers on the chest and a thin drawstring tie at the neckline, plus a pale sky-blue openwork crochet cardigan with elbow-length sleeves, clear mesh texture, and a light breathable material. Her accessories include one blue-green gradient feather earring on the left ear, bright cyan-blue at the top and deep teal at the bottom, about 5 cm long, and a matching thin cord choker with a small metal clasp. Her right hand gently supports her chin, and her left hand holds an artificial bouquet placed on her lap: seven blue five-petal flowers with slightly purple petal edges, deep blue star-shaped centers, green leaves, and several white buds. The background is a traditional tie-dye craft display. In the upper-left area hangs a white sign with black text: the first larger bold line means printing and tie-dye, and the second slightly smaller line means handmade craft; the text is sharp and horizontal. Behind and to the right are several hanging tie-dye fabrics on wooden racks: a dark indigo radial spiral cloth on the left, an orange-red cloth with white spiral tie-dye near the center-right, and a dark blue cloth with white concentric-circle and radial patterns on the right, all slightly wrinkled. The ground is gray cement steps and pavement. Sunlight comes diagonally from the upper right, casting clear rectangular light patches. Use natural daylight, soft contrast, a fresh artistic handmade-craft atmosphere, a medium portrait crop from waist to mid-thigh, the subject slightly left of center with empty space on the right, a standard non-distorted lens, moderate depth of field, and a blue-white-orange color palette with a fresh healing visual style. \\
\midrule
\texttt{2} &
Create a mind map or infographic about the book The Story of Art. At the top center, place the main title The Story of Art, with the English subtitle in parentheses below it, followed by a line explaining that it is a classic narrative of art history. The main body is a relationship network centered on a portrait of E.H. Gombrich, labeled with his name. Around him are four famous artists connected by lines. In the upper left is Leonardo da Vinci, represented by the Mona Lisa, labeled with his name and described as observing nature, pursuing perfection, and being a polymathic genius; the lines between him and Gombrich are labeled analysis and interpretation, and another line points to Raphael below with a tribute label. In the lower left is Raphael, represented by an angel image, labeled with his name and described as harmonious, elegant, ideal beauty, and a master painter; his connection to Gombrich is labeled tribute and inheritance. In the upper right is Michelangelo, represented by the head of David, labeled with his name and described as ambitious, shaping power, and master sculptor; his connection to Gombrich is labeled tribute and interpretation, and another line points down to Rembrandt with a tribute label. In the lower right is Rembrandt, represented by a self-portrait, labeled with his name and described as capturing the soul, master of light and shadow, and depicting humanity; his connection to Gombrich is labeled tribute and inheritance. Across the middle of the image, draw a left-to-right wavy timeline of art history: prehistoric art with a cave-painting bull icon and a note about cave paintings and statues; ancient Egypt and Mesopotamia with pyramid and sphinx icons and a note about eternity and order; Greece and Rome with a temple icon and a note about ideal and reality; the Middle Ages with a castle icon and a note about faith and symbolism; the Renaissance with a portrait icon and a note about humanism and rebirth; Baroque and Rococo with an ornamental badge icon and a note about drama and decoration; Neoclassicism and Romanticism with a profile icon and a note about reason and emotion; and Modern Art with an atom icon and a note about experimentation and abstraction. Below the timeline, add a centered sentence saying that art is not static but a constantly developing and changing story. At the bottom left, create a Core Themes section with five icons and labels: art-history primer, visual literacy, classic work, insight, and human spirit. At the bottom right, create a Character Cards section with two dark rectangular cards: one for Leonardo da Vinci with a side-profile silhouette and traits of curiosity, broad learning, and perfectionism, and one for Michelangelo with a side-profile silhouette and traits of determination, passion, and great ambition. \\
\bottomrule
\end{tabular}
}
\end{table}

\begin{table}[H]
\centering
\renewcommand{\arraystretch}{1}
\caption{\textbf{Unified evaluation template for Editing-Bench.} System prompt incorporating category-specific rubrics and scoring criteria for instruction-guided image editing.}
\label{tab:image_editing_prompt_template}
\resizebox{1.0\textwidth}{!}{
\begin{tabular}{p{\textwidth}}
\toprule
You are an expert evaluator for image editing results. You will receive one or more input/source images, the edit instruction, and one or more output image produced by an image editing model. Evaluate whether the output image correctly follows the requested edit while preserving the parts of the source image that should remain unchanged. Do not reward an image for being merely beautiful if it fails the edit. Do not penalize stylistic choices unless they conflict with the instruction or introduce visible artifacts. Use the full 1 to 5 scale. \\
\\
Return strict JSON only, with no markdown and no extra text. The JSON schema is: \\
\{"overall\_score": \textless{}float from 1 to 5 rounded to two decimals\textgreater{}, "sub\_scores": \{"criterion\_name": \textless{}float from 1 to 5\textgreater{}\}, "reasoning": "\textless{}one concise English sentence\textgreater{}", "failure\_modes": ["\textless{}short phrase\textgreater{}"]\} \\
\\
Score meanings: 5 = excellent edit with correct instruction following, source preservation, natural integration, and minimal artifacts; 4 = good with minor defects; 3 = partially correct but with clear issues; 2 = mostly failed or visibly broken; 1 = unusable, ignores the instruction, or severely corrupts the image. \\
\\
Category-specific rubric: \textcolor{red}{\texttt{\textless{}category-title\textgreater{}}} \\
\textcolor{red}{\texttt{\textless{}category-rubric\textgreater{}}} \\
\\
Sub-score criteria to include exactly with these keys: \\
\textcolor{red}{\texttt{\textless{}sub-score-criteria\textgreater{}}} \\
\\
The overall\_score should be a holistic score, not a simple arithmetic average. However, severe failure in instruction following, identity preservation, or text accuracy should strongly cap the overall score. \\
\bottomrule
\end{tabular}
}
\end{table}

\begin{table}[H]
\centering
\renewcommand{\arraystretch}{1.12}
\caption{\textbf{Category-specific evaluation rubrics for Editing-Bench.} Detailed values for \texttt{\textless{}category-title\textgreater{}} and \texttt{\textless{}category-rubric\textgreater{}} in the unified system prompt.}
\label{tab:image_editing_prompt_placeholders}
\resizebox{1.0\textwidth}{!}{
\begin{tabular}{p{0.24\textwidth}p{0.74\textwidth}}
\toprule
\texttt{\textless{}category-title\textgreater{}} & \texttt{\textless{}category-rubric\textgreater{}} \\
\midrule
Scene-level semantic transformation &
For this category, prioritize whether the whole scene transformation is semantically correct and physically coherent. A high score requires the edited scene to look like a single naturally captured image, not a pasted collage. \\
\midrule
Perceptual image enhancement &
For this category, do not reward hallucinated new content. Enhancement should improve clarity while keeping the source image faithful. \\
\midrule
Object-centric manipulation &
For deletion, the removed region should be plausibly filled. For addition/replacement, the new object must be recognizable and naturally integrated. \\
\midrule
Textual content editing &
For this category, text correctness and readability are critical. A visually pleasant image with wrong or unreadable text should receive a low score. \\
\midrule
Identity-preserving editing &
For this category, identity preservation is essential. Penalize face drift, identity change, unnatural anatomy, and over-editing of unrelated regions. \\
\midrule
Stylistic transfer &
For this category, style should change the appearance without destroying the source content or making the image structurally inconsistent. \\
\bottomrule
\end{tabular}
}
\end{table}

\begingroup
\small
\renewcommand{\arraystretch}{1.12}
\setlength{\LTleft}{\fill}
\setlength{\LTright}{\fill}

\begin{longtable}{
    @{}
    p{0.22\textwidth}
    p{\dimexpr0.73\textwidth-2\tabcolsep\relax}
    @{}
}
\caption{\textbf{Category-specific scoring criteria for Editing-Bench.} Detailed values for \texttt{\textless{}sub-score-criteria\textgreater{}} in the unified system prompt, capturing distinct quality requirements across editing tasks to support consistent and fine-grained assessment.}
\label{tab:image_editing_prompt_subscore_criteria}\\

\toprule
Category &
\texttt{\textless{}sub-score-criteria\textgreater{}} \\
\midrule
\endfirsthead

\multicolumn{2}{c}{\tablename~\thetable{} (continued)} \\
\toprule
Category &
\texttt{\textless{}sub-score-criteria\textgreater{}} \\
\midrule
\endhead

\midrule
\multicolumn{2}{r}{\footnotesize Continued on next page} \\
\endfoot

\bottomrule
\endlastfoot

Scene-level semantic transformation &
- Instruction following: Does the output implement the requested scene/background/composition transformation? \newline
- Source subject preservation: Are required people/objects/identity/clothing/details from the input preserved? \newline
- Global consistency: Are perspective, scale, lighting, shadows, and camera viewpoint coherent after the scene change? \newline
- Boundary integration: Are masks, edges, occlusion, and foreground-background transitions clean? \newline
- Visual quality: Is the final image free of distortions, texture melting, blur, noise, and generation artifacts? \\
\midrule

Perceptual image enhancement &
- Instruction following: Does the output perform the requested enhancement, restoration, or super-resolution operation? \newline
- Content preservation: Does it preserve the original content, identity, geometry, colors, and layout unless the instruction asks otherwise? \newline
- Detail recovery: Are details sharper and cleaner without hallucinated or over-smoothed structures? \newline
- Artifact control: Does it avoid ringing, oversharpening, waxy texture, noise amplification, and compression artifacts? \newline
- Visual quality: Is the enhanced output natural, high-quality, and technically clean? \\
\midrule

Object-centric manipulation &
- Instruction following: Is the requested object addition, deletion, replacement, or attribute modification correctly completed? \newline
- Target localization: Is the edit applied to the intended object/region without unintended changes elsewhere? \newline
- Source preservation: Are unrelated source content, identity, background, and composition preserved? \newline
- Physical integration: Do object scale, pose, lighting, shadows, contact, occlusion, and perspective fit the scene? \newline
- Visual quality: Are there no obvious seams, broken structure, duplicated parts, blur, or artifacts? \\
\midrule

Textual content editing &
- Instruction following: Does the output place, replace, remove, or modify the requested text exactly as instructed? \newline
- Text accuracy: Are spelling, characters, capitalization, punctuation, and requested wording correct? \newline
- Text readability: Is the text legible, sharp, correctly oriented, and not garbled or pseudo-text? \newline
- Layout style match: Does the text match the original layout, font style, perspective, material, lighting, and surface geometry? \newline
- Source preservation: Are non-text regions and unrelated details preserved? \newline
- Visual quality: Is the final image free of artifacts, blur, warped letters, and unnatural overlays? \\
\midrule

Identity-preserving editing &
- Instruction following: Does the output perform the requested clothing, pose, expression, hair, makeup, age, or face-related edit? \newline
- Identity preservation: Does the person keep the same recognizable identity and key facial/body characteristics where required? \newline
- Attribute accuracy: Are the requested changed attributes accurate and complete? \newline
- Anatomy and face quality: Are face, hands, limbs, pose, gaze, expression, and body proportions natural and undistorted? \newline
- Source preservation: Are unrelated clothing, background, accessories, and composition preserved unless the instruction changes them? \newline
- Visual quality: Is the final portrait technically clean, coherent, and artifact-free? \\
\midrule

Stylistic transfer &
- Instruction following: Does the output apply the requested style, color tone, filter, or artistic transformation? \newline
- Content preservation: Are the source subject, structure, layout, identity, and important details preserved? \newline
- Style consistency: Is the style/tone applied consistently across the image without patchy or conflicting regions? \newline
- Naturalness: Are lighting, contrast, saturation, texture, and material appearance coherent after the transfer? \newline
- Visual quality: Is the output free of over-processing, color banding, texture collapse, blur, and artifacts? \\

\end{longtable}
\endgroup

\section{Public Benchmark Evaluation}
\label{app:public_benchmarks}
This section evaluates \textbf{Qwen-Image-Flash} on Qwen-Image-Bench~\citep{li2026qwen} for T2I generation and GEditBench v2~\citep{jiang2026geditbench} for image editing. These public benchmarks complement our diagnostic evaluations by assessing the final unified model under independent evaluation protocols and contextualizing its performance against established models.

\paragraph{Results on Qwen-Image-Bench.}
Qwen-Image-Bench assesses five dimensions: image quality, aesthetics, prompt alignment, real-world fidelity, and creative generation. Table~\ref{tab:qwen_image_bench} compares \textbf{Qwen-Image-Flash} with representative generative models, with particular attention to Qwen-Image-2.0-Base and Qwen-Image-2.0-RL~\citep{xu2026qwen} as references for examining capability retention within the same model family.

With only $4$ NFEs, Qwen-Image-Flash achieves an overall score of $56.18$, exceeding the $80$-NFE Base model by $0.95$ points and outperforming FLUX 2 Max ($55.33$) and GPT Image 1 ($54.07$). Its gains over Base extend across all five dimensions, with the largest improvements in real-world fidelity ($+2.09$), prompt alignment ($+1.20$), and image quality ($+1.17$). This pattern indicates that the overall improvement is accompanied by gains in both semantic adherence and visual fidelity, without an observed regression in the remaining dimensions.

Relative to the RL reference, Qwen-Image-Flash trails by $1.66$ points overall, but the differences vary substantially across dimensions. Prompt alignment approaches the RL score ($58.84$ versus $59.28$), whereas creative generation exhibits the largest remaining gap ($59.23$ versus $64.94$). This uneven profile suggests that different aspects of post-trained performance are retained to different degrees under the current recipe, with creative generation leaving the greatest room for improvement. Overall, the public evaluation supports the effectiveness of the final model beyond our diagnostic benchmarks, while revealing capability-specific differences that are obscured by the aggregate score.

\begin{table*}[b]
\centering
\caption{\textbf{Comparison on Qwen-Image-Bench.} Results cover five complementary evaluation dimensions and the overall score. With only $4$ NFEs, Qwen-Image-Flash consistently surpasses the Base reference across all dimensions and achieves an overall score between the Base and RL references, demonstrating competitive generation performance under aggressive sampling compression. Higher scores indicate better performance.}
\label{tab:qwen_image_bench}
\small
\setlength{\tabcolsep}{5.0pt}
\renewcommand{\arraystretch}{1.08}

\begin{tabular}{lcccccc}
\toprule
\multirow{2}{*}{Model}
& \multicolumn{5}{c}{Dimension Score}
& \multirow{2}{*}{Overall $\uparrow$} \\
\cmidrule(lr){2-6}
& Quality $\uparrow$
& Aesth. $\uparrow$
& Align. $\uparrow$
& Fidelity $\uparrow$
& Creative $\uparrow$
& \\
\midrule

Nano Banana Pro
& 55.67
& 60.26
& 61.25
& 54.07
& 66.23
& 59.45 \\

FLUX 2 Max
& 53.64
& 56.85
& 57.35
& 49.35
& 56.50
& 55.33 \\

GPT Image 1
& 52.34
& 55.09
& 56.28
& 48.14
& 55.78
& 54.07 \\

\midrule

Qwen-Image-2.0-Base
& 52.29
& 57.10
& 57.64
& 47.54
& 58.22
& 55.23 \\

Qwen-Image-2.0-RL
& 54.39
& 58.67
& 59.28
& 51.83
& 64.94
& 57.84 \\

\textbf{Qwen-Image-Flash (Ours)}
& \textbf{53.46}
& \textbf{57.49}
& \textbf{58.84}
& \textbf{49.63}
& \textbf{59.23}
& \textbf{56.18} \\

\bottomrule
\end{tabular}
\end{table*}

\paragraph{Results on GEditBench v2.}
GEditBench v2~\citep{jiang2026geditbench} comprises $1{,}200$ real-world editing requests spanning $23$ tasks, extending evaluation to a broad range of practical editing scenarios. We assess instruction-guided editing through pairwise judgments from GPT-5.5, reporting Overall Elo, $95\%$ confidence intervals, and preference rates in Table~\ref{tab:geditbench_v2}.

With only $4$ NFEs, \textbf{Qwen-Image-Flash} achieves an Overall Elo of $1087$, closely matching FLUX.2 [klein] 9B ($1088$). Their similar preference rates ($0.6152$ and $0.6167$, respectively) further indicate comparable aggregate performance under this evaluation protocol. Qwen-Image-Flash also scores above Qwen-Image-2.0-Base ($1081$) and FLUX.2 [dev] Turbo ($1079$), although the overlapping confidence intervals warrant interpreting these differences as numerical advantages rather than conclusive improvements.

Within the Qwen-Image-2.0 family, the student falls between the Base and RL references, with a remaining gap of $25$ Elo points to the latter ($1112$). This positioning suggests that the current recipe sustains competitive editing performance under a restricted sampling budget, while leaving room to better retain the stronger teacher's capabilities. The same relative ordering on Qwen-Image-Bench provides complementary evidence that the final unified model remains competitive across both generation and editing under independent public evaluations.

\begin{table*}[t]
\centering
\caption{\textbf{Comparison on GEditBench v2.} Models are ranked by Overall Elo from deduplicated GPT-5.5 pairwise judgments, with $95\%$ confidence intervals (CI), preference rates (Pref.), and match counts reported. Higher Elo and preference rates indicate better performance.}
\label{tab:geditbench_v2}

\small
\setlength{\tabcolsep}{4.5pt}
\renewcommand{\arraystretch}{1.08}

\resizebox{\linewidth}{!}{%
\begin{tabular}{cllrrrrr}
\toprule
Rank
& Model
& Source
& Samples
& Overall Elo $\uparrow$
& 95\% CI
& Pref. $\uparrow$
& Matches \\
\midrule

1
& Nano Banana Pro
& Closed
& 1156
& 1218
& $-12/+14$
& 0.7703
& 20686 \\

3
& GPT Image 1.5
& Closed
& 1081
& 1095
& $-12/+13$
& 0.6246
& 19413 \\

4
& FLUX.2 [klein] 9B
& Open
& 1200
& 1088
& $-11/+11$
& 0.6167
& 21427 \\

7
& FLUX.2 [dev] Turbo
& Open
& 1200
& 1079
& $-11/+11$
& 0.6041
& 21427 \\

8
& FLUX.2 [klein] 4B
& Open
& 1200
& 1054
& $-10/+11$
& 0.5710
& 21427 \\

9
& Qwen-Image-Edit-2511
& Open
& 1200
& 1042
& $-9/+10$
& 0.5548
& 21427 \\

10
& Seedream 4.5
& Closed
& 1190
& 1036
& $-13/+13$
& 0.5473
& 21259 \\

11
& FLUX.2 [dev]
& Open
& 1200
& 1030
& $-11/+12$
& 0.5382
& 21427 \\

12
& LongCat-Image-Edit
& Closed
& 1200
& 1019
& $-10/+10$
& 0.5238
& 21427 \\

13
& Qwen-Image-Edit-2509
& Open
& 1200
& 986
& $-9/+9$
& 0.4782
& 21427 \\

14
& Qwen-Image-Edit
& Open
& 1200
& 960
& $-10/+10$
& 0.4439
& 21427 \\

15
& Step1X-Edit-v1p2
& Open
& 1200
& 890
& $-14/+13$
& 0.3517
& 21427 \\

16
& GLM-Image
& Open
& 1200
& 868
& $-12/+13$
& 0.3263
& 21427 \\

17
& BAGEL
& Open
& 1200
& 810
& $-13/+12$
& 0.2596
& 21427 \\

18
& FLUX.1 Kontext [dev]
& Open
& 1200
& 785
& $-12/+11$
& 0.2327
& 21427 \\

19
& OmniGen2
& Open
& 1200
& 760
& $-13/+13$
& 0.2088
& 21427 \\

\midrule

6
& Qwen-Image-2.0-Base
& Teacher
& 1200
& 1081
& $-10/+10$
& 0.6069
& 21427 \\

2
& Qwen-Image-2.0-RL
& Teacher
& 1200
& 1112
& $-11/+11$
& 0.6476
& 21427 \\

5
& \textbf{Qwen-Image-Flash}
& \textbf{Ours}
& 1200
& \textbf{1087}
& \textbf{$-9/+10$}
& \textbf{0.6152}
& 21427 \\

\bottomrule
\end{tabular}%
}
\end{table*}

\end{document}